\documentclass[letterpaper, 10 pt, conference]{ieeeconf}  

\IEEEoverridecommandlockouts                              

\usepackage{cite}
\usepackage{amsmath,amssymb,amsfonts}
\usepackage{algorithm}
\usepackage{algorithmic}
\usepackage{graphicx}
\usepackage{textcomp}
\usepackage{xcolor}
\usepackage{verbatim}

\let\labelindent\relax
\usepackage{enumitem}

\definecolor{Basak}{rgb}{0.18, 0.3, 0.7}

\usepackage[caption=false,font=footnotesize]{subfig}

\usepackage{acronym}

\acrodef{hmd}[HMD]{head-mounted display}
\acrodef{vr}[VR]{Virtual Reality}

\def\BibTeX{{\rm B\kern-.05em{\sc i\kern-.025em b}\kern-.08em
    T\kern-.1667em\lower.7ex\hbox{E}\kern-.125emX}}

\title{\LARGE \bf
TRAILS: Trajectory Generation for Automated Indoor Locomotion Studies
}

\author{Juho Kalliokoski$^{1}$, Evan G. Center$^{1}$, Steven M. LaValle$^{1}$, Timo Ojala$^{1}$, Ba\c{s}ak Sak\c{c}ak$^{1,2}$
\thanks{$^1$ Faculty of Information Technology and Electrical Engineering, University of Oulu, P.O. Box 4500, Oulu, FI-90014, Finland {\tt \footnotesize firstname.lastname@oulu.fi}}
\thanks{$^2$ Department of Advanced Computing Sciences, Maastricht University, Netherlands {\tt\footnotesize basak.sakcak@maastrichtuniversity.nl}}
}

\begin{document}

\maketitle
\thispagestyle{empty}
\pagestyle{empty}

\begin{abstract}
Virtual reality experiments involving navigation and simulated robot motion often require large, realistic environments and numerous distinct trajectories to avoid repetitive navigation while providing sufficient variation for controlled experiments. Designing such scenarios manually is labor-intensive, particularly when trajectories must satisfy predefined geometric and kinematic constraints while remaining diverse and comparable across experimental conditions. This paper presents TRAILS, an automated framework for generating complete virtual navigation scenarios consisting of an environment layout, a path, and a time-parameterized motion profile. The framework combines a constraint-based procedural environment generator inspired by Wave Function Collapse, trajectory generation using optimized cubic Bézier curves, and speed-profile generation subject to kinematic constraints. The generated scenarios provide diverse indoor environments with corridors, open spaces, and obstacles, together with trajectories of identical duration that vary in path geometry, wall proximity, and velocity profiles while minimizing systematic dependencies between these characteristics. 
The framework is designed as an easy-to-use tool requiring only a small number of intuitive user-defined parameters to automatically generate large sets of distinct navigation scenarios satisfying prescribed geometric and kinematic constraints. By relying on explicit, seed-controlled algorithmic procedures rather than data-driven models, the generated scenarios remain transparent, reproducible, and configurable, making the framework well suited for telepresence and human-subject experiments. The code is available at https://github.com/JucciK/Trails .

\end{abstract}

\section{INTRODUCTION}

Virtual Reality (VR), immersive telepresence, and human-robot interaction studies often require participants to experience controlled motion through virtual environments. In such experiments, the environment, the shape of traversed path, and the resulting motion may all influence navigation performance, presence, and cybersickness. Often, the used trajectories should be sufficiently diverse to avoid repeated exposure while remaining comparable with respect to duration and relevant motion characteristics.

Creating such scenarios by hand becomes increasingly difficult as the number and duration of needed trials increase. A researcher may need multiple distinct environments and trajectories while having to control properties such as path curvature, wall proximity, linear and angular velocity, and total exposure duration. Manually constructing and validating a large collection of environments and trajectories satisfying these requirements is time-consuming and can introduce unintended relationships between experimental variables.
Procedural generation offers a potential solution by enabling large numbers of unique environments and navigation scenarios to be produced automatically, but existing approaches generally focus on generating environment layouts rather than complete navigation scenarios consisting of realistic trajectories and motion profiles~\cite{hendrikx2013procedural, freiknecht2017survey, risi2020increasing}. 


This paper presents TRAILS, an automated framework for generating complete telepresence navigation scenarios in virtual environments. The proposed framework combines three stages: procedural environment generation, path generation and speed-profile generation. The environment-generation stage uses constraint-based tile expansion approach inspired by Wave Function Collapse~\cite{gumin2016wfc,karth2017wfc} to create large and varied indoor environments containing corridors, open spaces, and obstacles. A graph representation capturing the environment topology
is then constructed 
and converted into a path using constrained optimization of cubic Bézier curves~\cite{farin2002curves}, allowing path smoothness, curvature, and clearance requirements to be enforced. Finally, a speed profile is generated to satisfy constraints on linear and angular velocities while maintaining a predefined trajectory duration.

The primary goal of the framework is not to generate a single trajectory that is optimal with respect to a given objective function, but rather to automatically generate diverse navigation scenarios that satisfy predefined geometric and kinematic constraints.
By automating the generation process, the framework enables the researchers to evaluate different navigation methods under comparable conditions while systematically varying individual trajectory characteristics, such as curvature, wall proximity, and velocity profiles, without manually designing each scenario. In this way, the framework supports larger and more controlled experimental studies investigating user experience, cybersickness, and responses to different control methods for telepresence.


The remainder of this paper is organized as follows. Section~\ref{sec:rel_work} reviews the related literature. Section~\ref{sec:Trails} describes the individual components of the proposed TRAILS framework. Section~\ref{sec:num_example} provides example scenarios generated by TRAILS framework and evaluates various aspects of them. Finally, Section~\ref{sec:conc} summarizes the results and discusses limitations and future extensions.

\section{Related Work}\label{sec:rel_work}
\subsection{Procedural Environment Generation}
Procedural content generation (PCG) has been widely used to automatically generate virtual environments for games, simulations and interactive applications, reducing the need for manual design (see the surveys~\cite{freiknecht2017survey} and~\cite{hendrikx2013procedural} for a broad set of application areas of this method). 
The Wave Function Collapse (WFC) algorithm introduced by Gumin~\cite{gumin2016wfc} has gained popularity as a constraint-based approach for procedural environment generation. WFC algorithm generates content by enforcing local compatibility constraints between neighboring tiles, producing layouts that satisfy predefined structural rules while maintaining variability. Karth and Smith~\cite{karth2017wfc} later analyzed WFC as a form of constraint-solving and highlighted its suitability for procedural content generation tasks.
The environment-generation algorithm used in TRAILS follows a similar principle of local compatibility constraints but employs incremental tile-based expansion to ensure connectivity while allowing environments to be any size and shape.

More recent frameworks provide procedural generation at larger scales. ProcTHOR~\cite{deitke2022️} generates diverse interactive indoor environments with varying room layouts, materials, and object placement. Its primary goal is to provide large numbers of environments for training and evaluating embodied agents. Scenic~\cite{fremont2023scenic} approaches environment generation from different perspective. It provides a probabilistic programming language for describing distributions over scenes and agent behaviors. Such approaches demonstrate the usefulness of automatically generating diverse and controllable simulation scenarios.

\subsection{Path and Trajectory Generation}
Path planning is a fundamental problem in 
autonomous navigation, typically solved by searching an graph-based approximation of the robot's configuration space~\cite{lavalle1998rrt,pivtoraiko2009differentially}. In these representations, each edge in the graph is a kinematically feasible trajectory. Alternative to these methods, 
a path can be found in the
robot's workspace which is then smoothened for satisfying kinematic constraints. Bézier curves and B-splines are widely used for path-smoothing~\cite{simba2016real,benko2023minimum} because they can enforce continuity constraints and provide parametric representations suitable for finite-dimensional optimization.  



Temporal parameterization of paths has similarly been studied extensively in robotics. Velocity and acceleration profiles are often generated subject to constraints imposed by path curvature and vehicle dynamics~\cite{giannousakis2018mobile,bobrow1985time, tounsi1996trajectory}. Such methods commonly seek to minimize travel time, improve efficiency, or satisfy kinematic or dynamic constraints.


Controlled human-subject experiments often require distinct trajectories with comparable motion characteristics. Considering topological diversity, homotopy-aware planning methods exist that generate paths that are distinct in the way they avoid obstacles~\cite{sakcak2019homotopy,kuderer2014online,hernandez2015comparison}. However, topological diversity alone does not ensure the controlled variation required in these experiments. The present work therefore generates varied yet controlled speed profiles while maintaining a fixed trajectory duration and respecting predefined limits on linear and angular motion, allowing generated scenarios to remain comparable accross experimental conditions.

\subsection{Navigation and Locomotion Experiments}
VR research has extensively investigated how different locomotion methods influence navigation performance, spatial orientation, presence and cybersickness~\cite{hovrejvsi2025virtual,clifton2020effects,prithul2021teleportation}. These experiments rely on preconstructed environments and trajectories manually designed specifically for the study.
Some VR experiments use predefined trajectories as controlled motion stimuli: Riecke and Feuereissen~\cite{riecke2012move} varied path curvature to study vection, while Rouhani et al.~\cite{rouhani2024towards} varied velocity and vertical motion along a figure-eight path to benchmark VR sickness.


Similar considerations arise in immersive telepresence. Suomalainen et al.~\cite{suomalainen2021comfort} investigated how robot speed, turning, and proximity to walls and obstacles influence user comfort, while subsequent work showed that decoupling the user's viewing orientation from robot rotations improves comfort and reduces VR sickness~\cite{suomalainen2022unwinding}.
However, when experiments are conducted on a small number of predefined trajectories, individual characteristics of the trajectory may be confounded with the trajectories themselves. Broader stimulus sampling supports generalization beyond the tested examples~\cite{westfall2015replicating}, motivating automatic generation of diverse trajectories with systematically varied motion characteristics.

Whereas these studies provide valuable insights into user experience and motion perception in VR, the environments and navigation routes are typically created for the needs of a particular experiment. Relatively little attention has been given to 
automatically generating complete experimental scenarios consisting of a navigable environment, a realistic trajectory, and an associated motion profile. As experimental studies become larger and require greater environmental and trajectory diversity, automated generation methods can reduce the manual effort required to design and validate a sufficiently diverse set of trajectories and experimental scenarios. Controlling dependencies between different trajectory characteristics can help the investigation of their individual effects while maintaining comparable kinematic and geometric constraints across trajectories.


\section{TRAILS Framework}\label{sec:Trails}
\begin{figure*}[t]
    \centering
    \includegraphics[width=\textwidth]{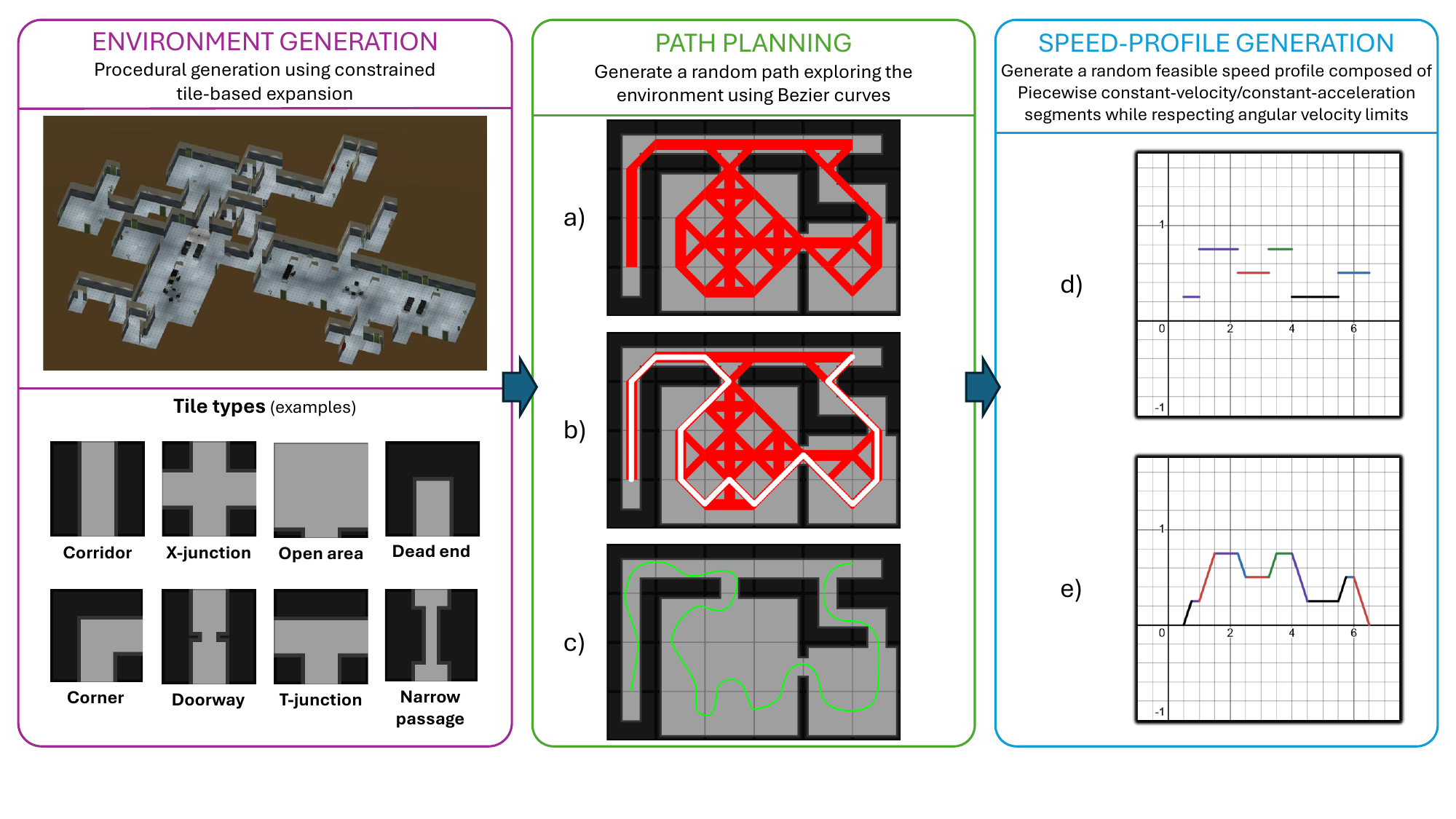}
    \vspace{-3.5em}
    \caption{Overview of the TRAILS framework.}
    \label{fig:pipeline}
\end{figure*}
In this section we detail each component of the proposed framework for generating complete scenarios of autonomous navigation in virtual environments. The objective is to automatically generate virtual environments and respective sets of distinct feasible trajectories with a given duration $T$ while limiting multicollinerity between relevant trajectory features (e.g., curvature, clearance from the obstacles and walls, velocity). 
The overall framework (illustrated in Figure~\ref{fig:pipeline}) consists of multiple stages, specifically, environment generation, path planning, and speed-profile generation. A final trajectory validation step is applied to discard trajectories that exceed a predefined variance inflation factor (VIF) threshold.




\subsection{Virtual Environment Generation}

Environment generation step of the framework is responsible for automatic generation of a virtual environment through repeated use of a set of given building blocks. The objective is to generate environments that satisfy the following requirements: 1) realistic appearance, 2) sufficient spatial extent to support trajectories with a given duration without repeating locations, 3) geometric variability (wide and narrow spaces), 4) obstacle presence, and 5) global path connectivity, that is, guaranteeing existence of a path connecting any point in the environment to any other.

\begin{figure}[t]
    \centering

    \subfloat[]{%
        \includegraphics[width=0.30\columnwidth]{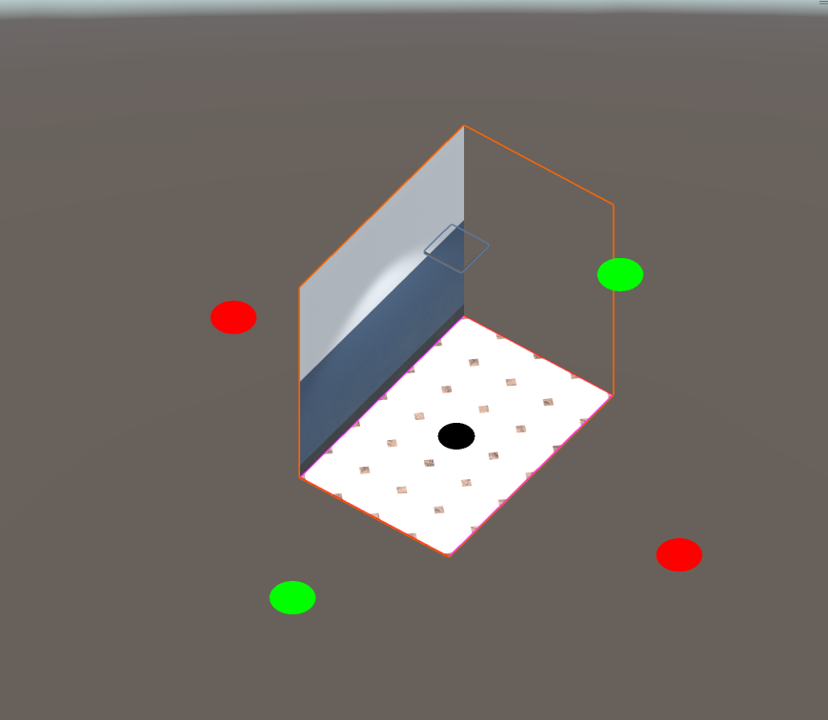}
        \label{fig:tile_labels}
    }
    \hfill
    \subfloat[]{%
        \includegraphics[width=0.30\columnwidth]{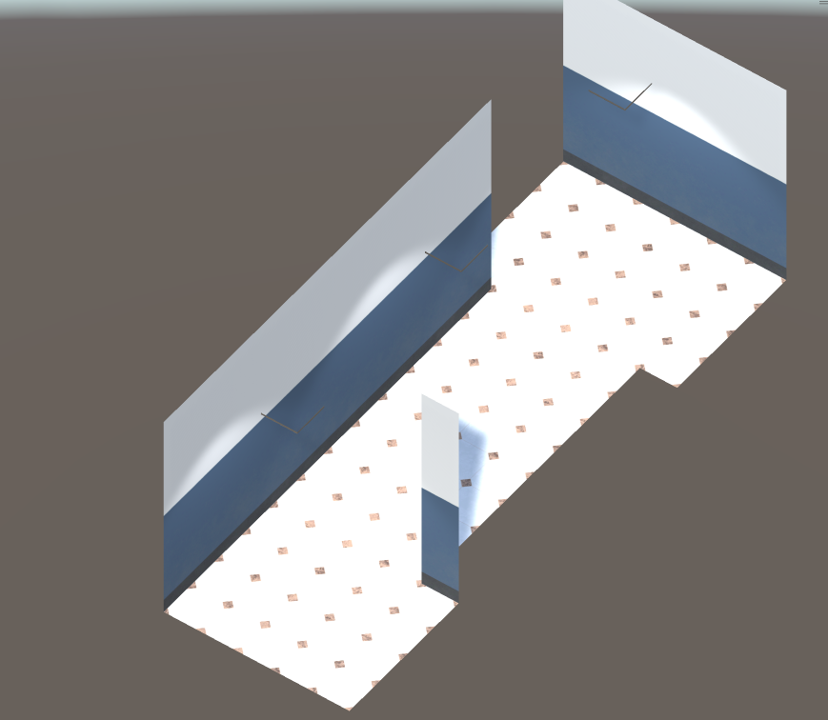}
        \label{fig:tile_valid}
    }
    \hfill
    \subfloat[]{%
        \includegraphics[width=0.30\columnwidth]{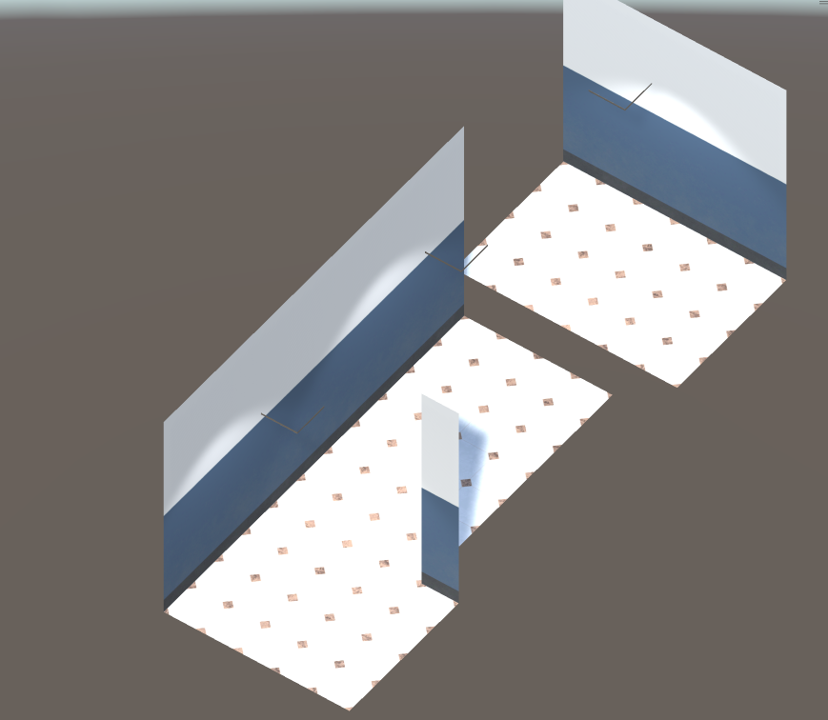}
        \label{fig:tile_invalid}
    }

    \caption{Tile adjacency constraints. (Left) Each side of a tile is assigned an interface label. A tile can be placed adjacent to another tile only if the labels on the shared edge are compatible. (Middle) Compatible neighboring tiles satisfying adjacency constraints. (Right) Incompatible neighboring tiles.}
    \label{fig:tile_constraints}
\end{figure}
The environment is assembled using a finite set $\mathcal{T}$ of modular tiles (pieces). Each tile is a three-dimensional cube and can be connected to the environment by placing it on a regular planar grid $\mathcal{G} \subseteq \mathbb{Z}\times \mathbb{Z}$ while respecting the local adjacency constraints. These constraints are defined by interface labels assigned to each side of the tile, indicating whether the side corresponds to a corridor opening, hall opening, hall opening with side wall, doorway, or closed wall. Each tile is represented by a tuple $\tau$ of interface labels
    $\tau= (l_N, l_E, l_S, l_W),$
where $l_N, l_E, l_S, l_W$ denote the interface labels assigned to the north, east, south, and west sides of the tile, respectively (see Figure~\ref{fig:tile_constraints}\;left). A tile $\tau$ is admissible at location $p$ if and only if for every occupied neighbor the shared sides with the adjacent tiles satisfy pairwise compatibility constraints. For example, placing a corridor segment enforces that adjacent cells along the corridor's sides cannot contain pieces with openings facing the corridor walls, while the cell directly in front of the corridor must contain a compatible corridor or junction piece (see Figure~\ref{fig:tile_constraints} middle and right). 

Procedural generation of the environment is inspired by the Wave Function Collapse (WFC) algorithm~\cite{gumin2016wfc}. However, our approach differs from WFC in two aspects to ensure that global connectivity requirement is satisfied. First, instead of maintaining a global superposition over all grid cells, the environment is generated incrementally through frontier expansion. Second, connectivity is enforced constructively during generation rather than verified afterwards. Note that whereas in WFC each cell of a regular grid contains a tile, it is possible in our approach that certain cells contain no tiles (brown regions in the left block of Figure~\ref{fig:pipeline}). 

Environment generation (Algorithm~\ref{alg:env-gen}) follows an iterative expansion strategy from the environment frontier. The set of grid positions corresponding to unexpanded valid interfaces of already placed tiles define the frontier set $F \subseteq \mathcal{G}$. The algorithm starts with placing a seed tile $s$ at a user defined position $p_0 \in \mathcal{G}$ on the empty grid $\mathcal{G}$. Consecutively, the grid positions corresponding to the open sides of $s$ are inserted to $F$. At each iteration, a grid position $p \in F$ is selected, and the set of feasible tiles $S \subseteq \mathcal{T}$ is determined by keeping only those tiles whose interface labels satisfy the adjacency constraints imposed by the neighboring tiles already placed in the grid. A tile is then sampled from S according to a weighted probability distribution, where each tile type is assigned a predefined weight reflecting its desired frequency in the generated environment. For example, corridor tiles are assigned higher weights than doorway tiles to avoid excessive numbers of doorways while preserving variability. The sampled tile is then placed at position $p$.
Because each newly added tile is connected to the existing structure through a valid interface, the resulting environment remains a single connected component by generation. Consequently, global connectivity is preserved, ensuring that any two points in the environment are mutually reachable via a valid path. 
Every time a new tile $\tau$ is inserted into $\mathcal{G}$, the frontier set $F$ is updated to remove the processed frontier $p$ and add new elements to $F$ induced by the newly placed tile. 

Tiles are sampled without any bias until the number of placed tiles reaches a predefined environment size $N_{\min}$. Once $N_{\min}$ is reached, sampling is modified to favor tiles that reduce expansion of the frontier to ensure closure. In this phase, the selection strategy is modified to reduce the number of remaining open connections in the environment. Each candidate tile $\tau \in \mathcal{T}$ is assigned a score based on the number of new openings it introduces when placed at a given position. Let $o(\tau)$ denote the number of openings created by placing tile $\tau$. During the closure phase, each feasible tile ($\tau \in S$) is evaluated by temporarily placing it at the candidate position and counting the resulting number of open sides. The minimum value is then computed as $o_{\min}=\min_{\tau\in S} o(\tau)$. Only tiles satisfying $o(\tau)=o_{\min}$ are retained for sampling. The weighted random selection is then applied only to this reduced candidate set. Thus, the closure phase explicitly excludes feasible tiles that would leave more open connections than the best available alternatives.

In addition to the structural layout, static obstacles are generated immediately after each tile is placed. Each tile defines a set of candidate obstacles, each associated with a predefined placement probability, and each candidate is placed independently according to a Bernoulli trial. Obstacle placement is constrained to the footprint of the piece and does not affect connectivity between pieces. This introduces visual and navigational variability without altering the underlying structure of the environment.

\begin{algorithm}[t]
\caption{Virtual Environment Generation}
\label{alg:env-gen}
\begin{algorithmic}[1]

\REQUIRE Tile set $\mathcal{T}$ with constraints and weights, seed tile $s$, minimum size $N_{\min}$
\ENSURE Generated environment $\mathcal{G}$

\REPEAT
    \STATE Initialize empty grid $\mathcal{G}$
    \STATE Initialize empty frontier list $F$
    \STATE placed $\leftarrow 0$

    \STATE Place seed tile $s$ at user-defined position $p_0$
    \STATE Update constraints around $p_0$
    \STATE Add valid openings of $s$ to $F$
    \STATE placed $\leftarrow 1$
    \STATE success $\leftarrow$ true

    \WHILE{$F \neq \emptyset$}
        \STATE Pop first position $p$ from $F$

        \STATE $S \leftarrow \{\tau \in \mathcal{T} \mid t \text{ satisfies constraints at } p\}$

        \IF{$S = \emptyset$}
            \STATE Restart generation
        \ENDIF

        \IF{placed $\geq N_{\min}$}
            \STATE Filter $S$ to prefer tiles that reduce future openings
        \ENDIF

        \STATE Sample tile $\tau$ from $S$ using weighted random selection
        \STATE Place $\tau$ at position $p$ in $\mathcal{G}$
        \STATE placed $\leftarrow$ placed + 1

        \STATE Update constraints around $p$
        \STATE Add new valid openings to $F$
    \ENDWHILE

\UNTIL{success = true}

\RETURN $\mathcal{G}$

\end{algorithmic}
\end{algorithm}

\subsection{Path Generation}
For path and trajectory generation, we consider a telepresence robot with kinematics equivalent to a unicycle given by 
\begin{equation}
\label{eqn:unicycle}
\begin{alignedat}{2}
\dot{x}&=v\cos\theta \\
\dot{y}&=v\sin\theta \\
\dot{\theta}&=\omega,
\end{alignedat}
\end{equation}
where $(x,y)$ denotes the robot position, $\theta$ its heading, and $v$ and $\omega$ the linear and angular velocities, respectively. The input $u = (v,\omega)$ is subject to constraints and takes part in a compact set $U \subseteq \mathbb{R}^2$. Let $Q \subseteq W \times S^1$, in which $W$ is the planar environment the robot moves on and $S^1$ denotes the unit circle. There are configurations $q=(x,y,\theta) \in Q$ that the robot is in collision with the environment boundaries or the obstacles given its geometry. The set of configurations that are free from collisions is denoted by $Q_{\text{free}}$. A trajectory $\sigma : [0,T]\rightarrow Q$ is called feasible if $\sigma(t)$ satisfies Equation~\eqref{eqn:unicycle} and that $\sigma(t) \in Q_{\text{free}}$ for all $t \in [0,T]$.
The particular robot motion model is chosen since it is consistent with commercial telepresence robots~\cite{double, gobe} as well as models used in research for simulation~\cite{das2025adaptable} and real-world experiments~\cite{jones2020vroom,239144,beno2018work}. 

Whereas all trajectories are required to have the same given duration $T$, it is desired that they vary in their features. Specifically, we consider the following features: closest distance to the walls, linear and angular velocity profiles, and the path curvature. To enable independent analysis in subsequent user studies, the trajectories should promote variation in these features while limiting multicollinearity between them. 
To ensure variability in features, trajectory generation is split into two parts: path generation and speed profile generation.
Path generation (middle block in Figure~\ref{fig:pipeline}) step deals with generating a smooth path that is collision free. This step starts from a polygonal representation of the robot workspace (plane that the robot moves on) corresponding to the navigable region and determines a route which is then smoothed to obtain a path with desired characteristics. 

\subsubsection{Polygonal Representation}
Let $O \subseteq W$, in which $W$ is the plane that the robot moves on, be the set of obstacle regions, including walls and static obstacles. The navigable workspace is therefore given by $W\setminus O$. The navigable space is partitioned into a set of 
polygons.
An adjacency graph $G=(V,E)$, in which $V$ is the set of vertices corresponding to the traversable polygon interfaces (openings) and $E$ is the set of edges. 
Two vertices are connected by an edge whenever the corresponding openings can be traversed consecutively through a common polygon. A path in the graph represents a sequence of openings through which the robot travels while traversing the corresponding polygons.

\begin{algorithm}[t]
\caption{Coarse Path Generation on the Polygon Graph}
\label{alg:path_generation}
\begin{algorithmic}[1]

\REQUIRE Polygon decomposition, minimum coarse path length $L$
\ENSURE Coarse path $P$

\STATE Construct an opening graph $G=(V,E)$
\STATE Each vertex $v\in V$ corresponds to a traversable polygon edge/opening
\STATE Add $(v_i,v_j)\in E$ if openings $v_i$ and $v_j$ are connected through a common traversable polygon

\STATE Select a random start opening $v_0 \in V$
\STATE $P \leftarrow [v_0]$
\STATE $V_{\mathrm{visited}} \leftarrow \{v_0\}$
\STATE $v \leftarrow v_0$

\WHILE{$|P| < L$}
    \STATE $N \leftarrow \{u \mid (v,u)\in E,\; u \notin V_{\mathrm{visited}}\}$
    \IF{$N=\emptyset$}
        \STATE \textbf{break}
    \ENDIF
    \STATE Select $u$ uniformly at random from $N$
    \STATE Append $u$ to $P$
    \STATE $V_{\mathrm{visited}} \leftarrow V_{\mathrm{visited}} \cup \{u\}$
    \STATE $v \leftarrow u$
\ENDWHILE

\STATE \textbf{return} $P$

\end{algorithmic}
\end{algorithm}

Coarse path generation (Algorithm~\ref{alg:path_generation}) begins by constructing a 
path on the polygon adjacency graph. Starting from an initial polygon, the algorithm repeatedly selects one of the adjacent, previously unvisited polygons and extends the path to that polygon. To encourage exploration of the environment, polygons already included in the path are excluded from further consideration, ensuring that each polygon is visited at most once. This prevents loops within a polygon and encourages the path to traverse distinct regions of the environment.

The process is repeated until a predefined minimum path length is reached. The resulting coarse path $P$ is therefore represented as an ordered sequence of adjacent polygons, which subsequently serves as the basis for the continuous trajectory generation described in the following section. An example of resulting coarse 
path is shown in Figure~\ref{fig:pipeline}~b.

\subsubsection{Path Generation}
The coarse polygon path defines only the 
route through the environment and therefore is not smooth.
To obtain a smooth 
path trackable by non-holonomic mobile robots, the coarse path is converted into a spline consisting of cubic Bézier curve segments. A cubic Bézier curve is defined by
\begin{equation}
\mathbf{B}(\mu)
=
(1-\mu)^3\mathbf{P}_0
+
3(1-\mu)^2\mu\,\mathbf{P}_1
+
3(1-\mu)\mu^2\,\mathbf{P}_2
+
\mu^3\mathbf{P}_3,
\end{equation}
where $\mathbf{P}_0, \mathbf{P}_3$ denote the endpoints of the curve, $\mathbf{P}_1$ and $\mathbf{P}_2$ are the control points, and $\mu \in [0,1]$ is the curve parameter.

The coarse path $P$ is partitioned into segments consisting of a fixed number of consecutive polygon edges. The Bézier curves comprising the spline are generated sequentially across segments. However, all Bézier curves within a segment are optimized simultaneously, while continuity between neighboring segments is maintained by fixing the starting point and first control point of each segment to the optimized coefficients of the previous segment. 

For each segment, the optimization variables are the remaining Bézier endpoints and control points. Each endpoint is constrained to lie on its corresponding polygon edge and may be displaced from the edge midpoint by up to a fraction $\alpha$ of the edge length, where $0 \leq \alpha \leq \alpha_{\max}$,  along the edge. This allows the trajectory to vary within corridors and open spaces while ensuring that endpoints remain inside the associated polygons. Each control point is parametrized by its distance $d$ and angle $\phi$ relative to its endpoint, with $|\phi| \leq \phi_{\max}$ and $d_{\min} \leq d \leq d_{\max}$, where $\phi_{\max}, d_{\min}$, and $d_{\max}$ denote the respective limits.


The free parameters of Bézier curves are optimized subject to previously described constraints using the COBYLA algorithm~\cite{powell1994direct}. Paths are evaluated by uniformly sampling each Bézier curve at $N_s$ points and minimizing the following objective function
\begin{equation}
\begin{aligned}
J =\;&
w_o N_o
+w_\omega N_\omega
+w_d N_d \
+\lambda_c C
+\lambda_l L ,
\end{aligned}
\end{equation}
where $N_o$ denotes the number of sampled points lying outside the floor polygon associated with the Bézier curve on which they were sampled, $N_d$ the number of sampled points lying closer than the prescribed minimum clearance $d_{\mathrm{wall}}$ to the nearest wall or obstacle, $N_\omega$ the number of sampled points for which the estimated angular velocity exceeds the prescribed limit, $C$ 
the sum of sampled absolute curvatures,
and $L$ the path-length penalty. The coefficients $w_o$, $w_\omega$, and $w_d$ are large penalty weights enforcing geometric and kinematic feasibility, whereas $\lambda_c$ and $\lambda_l$ control the influence of the curvature and path-length objectives respectively.

Angular velocity at each sampled point is estimated from the local path curvature according to
\begin{equation}
    \omega_i=\kappa_i v_n,
\end{equation}
where $\kappa_i$ denotes the curvature at the $i$th sampled point and $v_n$ is the nominal forward velocity used solely for evaluating kinematic feasibility. Sampled points satisfying $|\omega_i|>\omega_{\max}$ contribute to $N_\omega$.
The final two terms of the objective function determine the geometric style of the generated path. The curvature objective is defined as
\begin{equation}
C=\sum_{i=1}^{N_s}\left|\kappa_i\right|.
\end{equation}

Each path segment is randomly assigned to either a \emph{straight} or a \emph{windy} category before optimization. For straight segments, $\lambda_c>0$, encouraging trajectories with lower cumulative curvature, whereas for windy segments, $\lambda_c<0$, encouraging trajectories with higher cumulative curvature while still satisfying the feasibility constraints.

Rather than directly minimizing the path length, a target length $l_t$ is computed from the corresponding coarse path segment. The path-length penalty is defined as

\begin{equation}
L=(l-l_t)^2,
\end{equation}
where $l$ denotes the length of the optimized Bézier path. The target length $l_t$ is obtained by adding or subtracting a predefined offset from the corresponding coarse path length depending on whether the segment is designated as straight or windy. Consequently, straighter segments are encouraged to follow shorter paths, whereas windy segments are encouraged to traverse longer paths while remaining consistent with the underlying coarse route.

\subsection{Speed Profile Generation}
After spatial paths are generated, a time-parameterized speed profile is computed for each path to control the robot's linear and angular motion over time. The objectives of this step are to ensure identical trajectory duration across participants, enforce a fixed average forward velocity, and respect constraints on angular velocity and acceleration derived from path curvature.
All trajectories are assigned a fixed duration of $T$ seconds, and a target average linear velocity $v_{avg}$. This ensured that all participants experienced motion of equal duration and comparable overall motion intensity.

\subsubsection{Candidate Speed and Acceleration Segments}
Speed profiles are constructed from a predefined set of constant-speed segments with target velocities ranging from $v_{\min}$ to $v_{\max}$. Transitions between target velocities are achieved using constant acceleration segments, with allowable accelerations and decelerations ranging from $\pm a_{\min}$ to $\pm a_{\max}$.

The total trajectory time is divided into a fixed number of speed segments, with $N_{seg}$ segments allocated for each possible target velocity (Fig.~\ref{fig:pipeline} d). To introduce variability in segment duration while preserving the total time spent at each velocity, half of the segments are shortened, while the remaining half are lengthened by the same amounts. For each shortened segment the reduction in duration is sampled uniformly from the interval $\Delta t_s \sim \mathcal{U}(0,1)$, where $t_s$ is the original segment duration. The same amoung, $\Delta t_s$, is then added to one of the remaining segments, ensuring that the total time spent at each target velocity remains unchanged. This procedure preserves a constant total time per velocity while producing non-uniform segment lengths.

\subsubsection{Construction of the Speed Profile}
Speed profile generation proceeds incrementally, starting from zero velocity. At each step, candidate combinations of remaining speed segments and acceleration values are evaluated. For each candidate, and acceleration ramp is constructed to transition from the current velocity to the candidate segment's target velocity, followed by the constant-speed segment itself.

Each candidate segment sequence is validated against the path using the underlying Bézier path representation. Validation is performed by simulating the motion along the path and checking kinematic constraints at discrete points. In particular, instantaneous angular velocity is estimated as the product of local path curvature and linear speed, and angular acceleration is derived from changes in curvature and speed. Candidate segments are rejected if the angular velocity exceeded a predefined maximum angular velocity, $\omega_{\max}$, or if the angular acceleration exceeded a predefined maximum angular acceleration, $\dot{\omega}_{\max}$.
From the set of valid candidates, one segment is selected uniformly at random and appended to the speed profile. This process is repeated until all predefined speed segments had been placed. The final segment is adjusted to decelerate the robot smoothly back to zero velocity.

\subsubsection{Average Speed Adjustment}
After constructing a complete speed profile, the resulting average velocity is computed. If the average velocity deviated from the target value, a correction step is applied. This adjustment modifies the durations of two constant-speed segments: the longest segment at the highest velocity and the shortest segment at the lowest velocity (or vice versa, depending on the direction of correction). Segment durations are adjusted analytically to match the target average velocity while preserving total trajectory duration.
If there are no valid adjustments (e.g., due to negative segment durations), the entire speed profile generation process is repeated. This retry mechanism ensures that only valid speed profiles were retained.

The resulting output after these steps is a complete navigation scenario consisting of an environment, a spatial trajectory, and an associated motion profile suitable for use in virtual reality experiments. An example of final speed profile can be seen in Fig.~\ref{fig:pipeline} e.


\subsection{Trajectory Validation}

After both the path and speed profile have been generated, trajectory characteristics including curvature, wall clearance, and linear and angular velocity are sampled along the trajectory. Variance inflation factors (VIFs) are calculated to identify multicollinearity between these characteristics. Trajectories exceeding predefined VIF threshold are discarded and regenerated, reducing potential confounding between the characteristics. VIFs are evaluated separately within each trajectory. Dependencies across the set of trajectories are not considered in the implementation, though such dependencies would only be problematic for between-subjects rather than within-subjects measures. Validation could be altered to include evaluation the full set of trajectories and regeneration of trajectories when needed, but this would increase the computational cost significantly.

\section{Numerical Example}\label{sec:num_example}
\begin{figure}[t]
    \centering
    \includegraphics[width=\columnwidth]{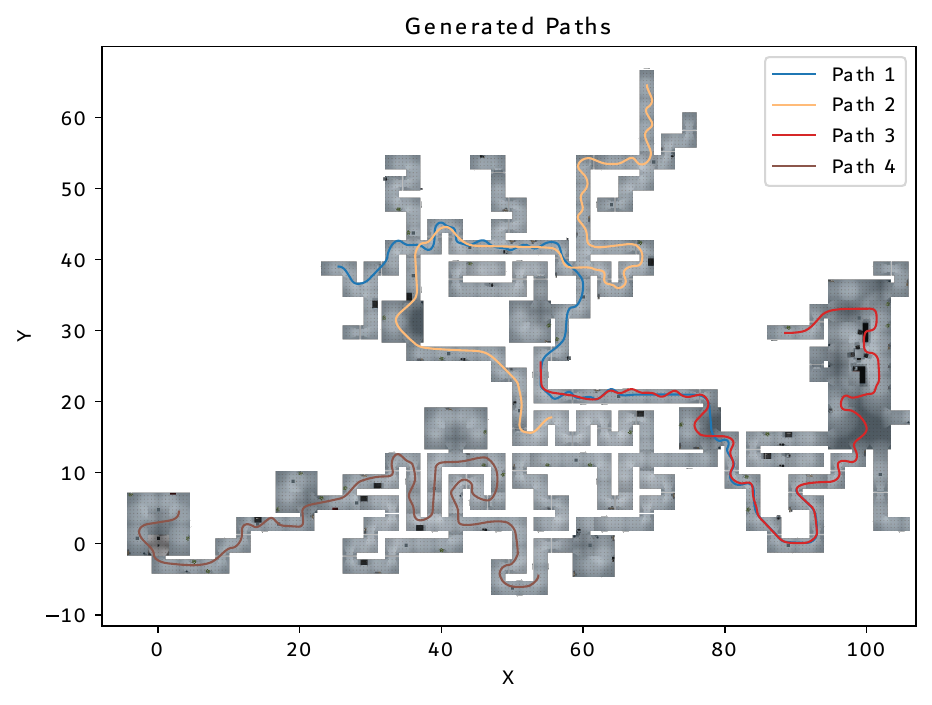}
     \vspace{-2em}
    \caption{Example environment (top-view) generated by the proposed framework together with four generated paths.}
    \label{fig:example_paths}
      \vspace{-1em}
\end{figure}

\begin{figure}[t]
\centering

\subfloat[]{%
    \includegraphics[width=0.48\columnwidth]{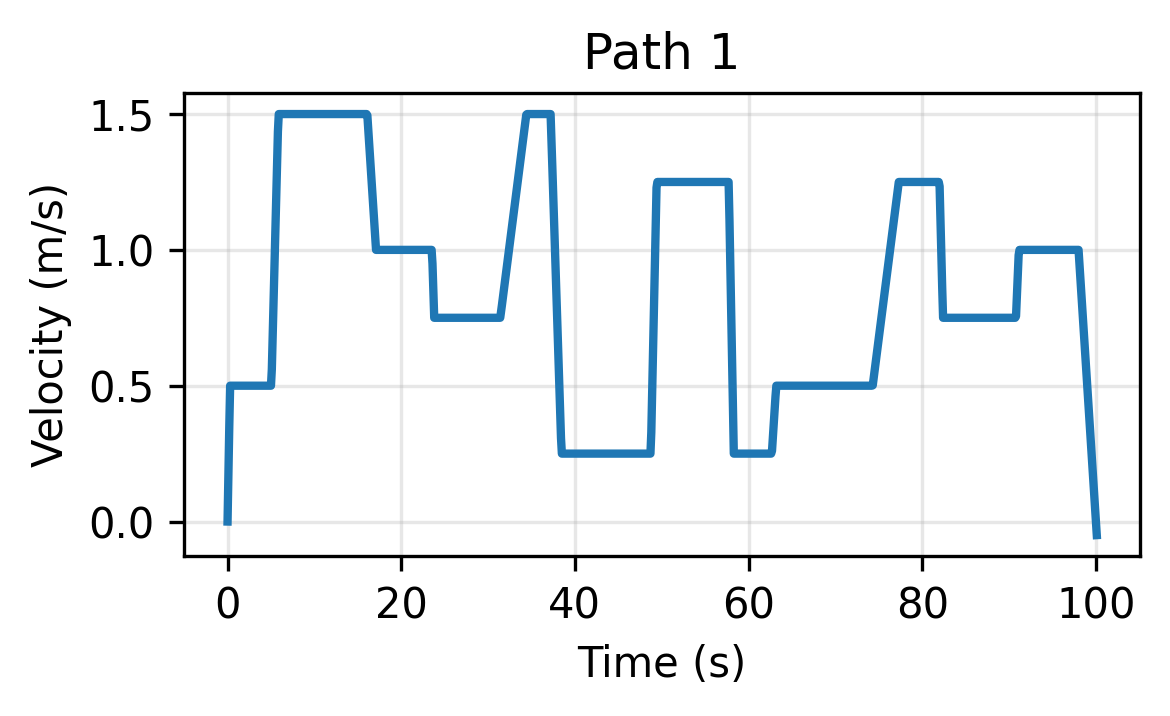}
}
\hfill
\subfloat[]{%
    \includegraphics[width=0.48\columnwidth]{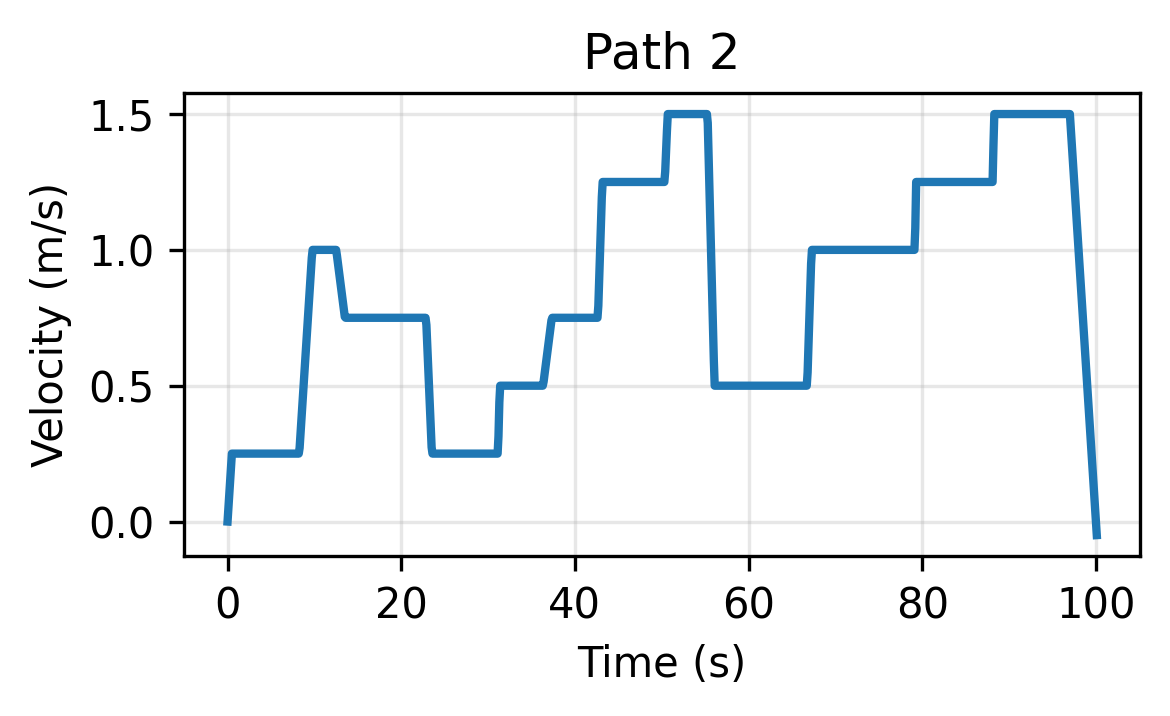}
}

\vspace{0.3cm}

\subfloat[]{%
    \includegraphics[width=0.48\columnwidth]{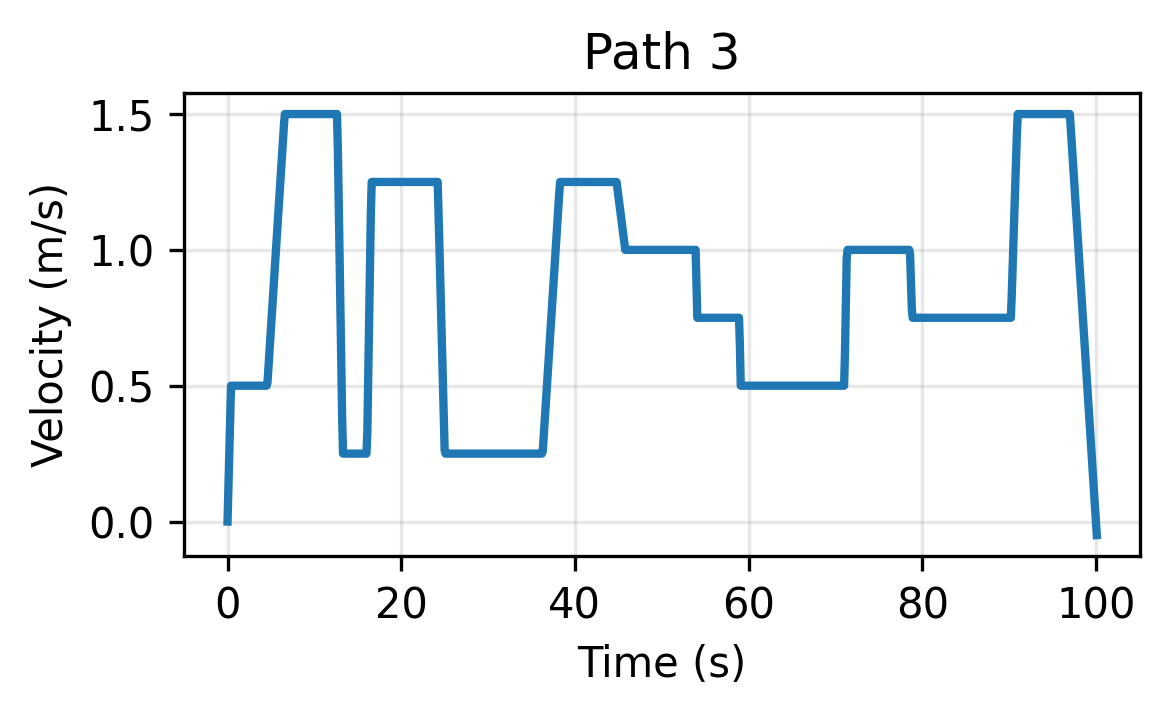}
}
\hfill
\subfloat[]{%
    \includegraphics[width=0.48\columnwidth]{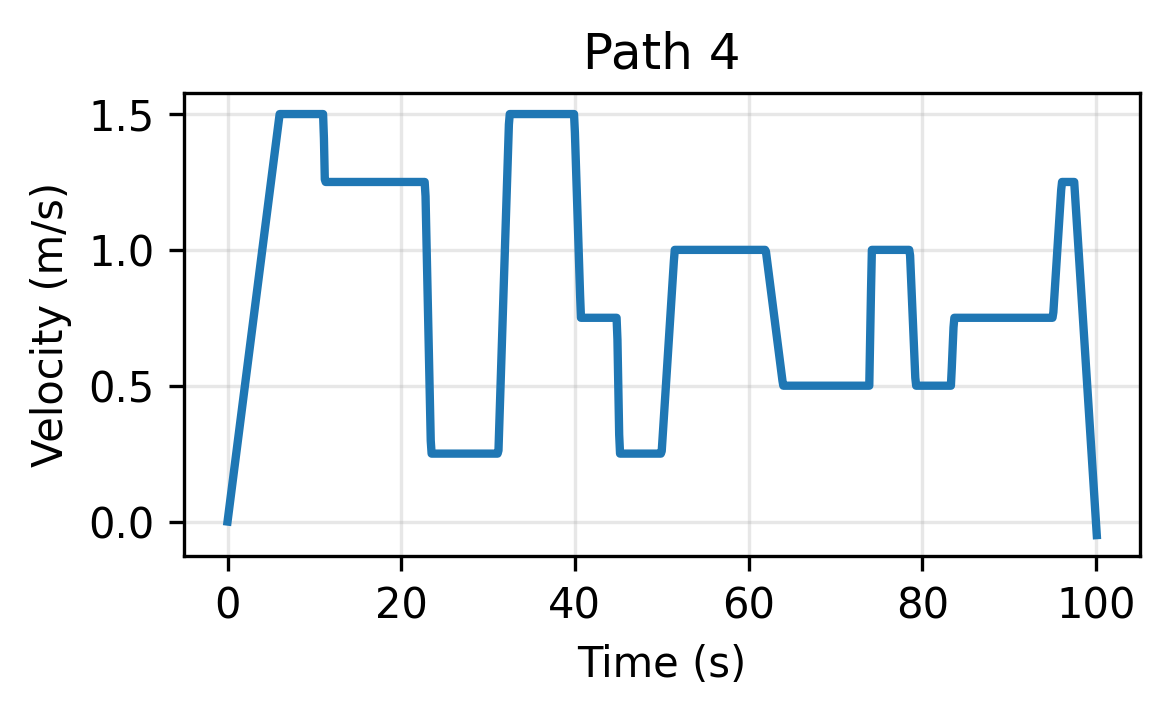}
}

\caption{Linear velocity profiles corresponding to the trajectories shown in Fig.~\ref{fig:example_paths}. The profiles consist of piecewise constant-velocity segments connected by constant-acceleration transitions while satisfying the prescribed motion constraints.}
\label{fig:example_speed}
\end{figure}

To illustrate the proposed framework, an example navigation scenario is generated using the parameter values listed in Table~\ref{tab:parameters}.

Figure~\ref{fig:example_paths} shows one environment generated by Algorithm~\ref{alg:env-gen} together with four generated paths. Although all paths satisfy the same constraints and have identical durations, they exhibit different characteristics, including variations in curvature, wall proximity, and traversal through narrow and open regions.

Example speed profiles corresponding to the generated trajectories are shown in Fig.~\ref{fig:example_speed}. Each profile consists of piecewise constant-velocity and constant-acceleration segments that satisfy the prescribed limits on linear and angular motion while maintaining the desired traversal time.


\begin{table}[t]
\centering
\caption{Parameters used in the numerical example.}
\label{tab:parameters}
\begin{tabular}{lc}
\hline
Parameter & Value\\
\hline
Minimum number of tiles & 500\\
Maximum endpoint offset $\alpha_{\max}$ & 0.4\\
Maximum control-point angle $\phi_{\max}$ & 1.0 rad\\
Control-point distance $[d_{\min},d_{\max}]$ & $[0.25,1.5]$ m\\
Minimum wall clearance $d_{\mathrm{wall}}$ & 0.30 m\\
Maximum angular velocity $\omega_{\max}$ & $\pi$ rad/s\\
Target trajectory duration & 100 s\\
Average velocity & 1.0 m/s\\
VIF threshold $\mathrm{VIF}_{\max}$ & 2.5\\
\hline
\end{tabular}
\end{table}

To evaluate the computational cost of the trajectory generation, 10 trajectories were generated with a duration of $100~s$ and 10 trajectories with a duration of $200~s$. The total time required to obtain each valid trajectory is measured, including unsuccessful generation attempts that were subsequently discarded and regenerated. The mean generation time is $85.97 \pm 23.15~s$ for $100~s$ trajectories and $416.57 \pm 44.47~s$ for $200~s$ trajectories (mean $\pm$ standard deviation).
\section{Discussion and Conclusion}\label{sec:conc}

This paper presents TRAILS, an automated framework for generating complete navigation scenarios for virtual reality and telepresence experiments. Unlike previous work that typically addresses environment generation, path planning, or motion generation as separate problems, the proposed framework integrates these components into a single pipeline capable of producing connected indoor environments, smooth robot trajectories, and corresponding time-parameterized motion profiles. The resulting scenarios satisfy predefined geometric and kinematic constraints while providing controlled variation in path geometry, wall proximity, and robot motion characteristics.

The framework is developed to reduce the effort required to create large numbers of experimental scenarios. By automatically generating environments together with feasible trajectories and speed profiles, the framework enables researchers to create diverse navigation scenarios that remain comparable across experimental conditions. Although motivated by immersive telepresence research, the generated scenarios may also be applicable to other virtual reality studies involving navigation, human-robot interaction, and mobile robot simulation.

The proposed framework still has several limitations. The environment generator relies on a predefined library of modular environment pieces, limiting the variety of generated layouts to the available tile set. The tiles used in this work form the default tile set supplied with the framework. However, the generation algorithm is not limited to these tiles, and users may define additional tiles by themselves. Likewise, trajectory generation is based on heuristic optimization and predefined parameter ranges rather than global optimization. The generated scenarios are also intended for simulated environments and do not explicitly account for uncertainties encountered in real-world robot navigation. Finally, validation using VIF thresholding only assures against multicollinearity, while potential non-linear relationships between variables would remain hidden. 

Although the framework targets non-holonomic mobile robots, its modular structure makes it applicable to a broader range of navigation systems. Environment generation is independent of the robot model, while the path generation and speed-profile generation stages can be adapted by replacing the underlying motion model, kinematic constraints, and optimization objectives. Consequently the proposed framework could be extended to support holonomic mobile robots, aerial vehicles, or other robotic platforms requiring different motion characteristics.

\addtolength{\textheight}{-12cm}   









\bibliographystyle{IEEEtran}
\bibliography{bib}

@article{freiknecht2017survey,
  title={A survey on the procedural generation of virtual worlds},
  author={Freiknecht, Jonas and Effelsberg, Wolfgang},
  journal={Multimodal Technologies and Interaction},
  volume={1},
  number={4},
  pages={27},
  year={2017},
  publisher={MDPI}
}

@article{hendrikx2013procedural,
  title={Procedural content generation for games: A survey},
  author={Hendrikx, Mark and Meijer, Sebastiaan and Van Der Velden, Joeri and Iosup, Alexandru},
  journal={ACM Transactions on Multimedia Computing, Communications, and Applications (TOMM)},
  volume={9},
  number={1},
  pages={1--22},
  year={2013},
  publisher={ACM New York, NY, USA}
}

@article{risi2020increasing,
  title={Increasing generality in machine learning through procedural content generation},
  author={Risi, Sebastian and Togelius, Julian},
  journal={Nature Machine Intelligence},
  volume={2},
  number={8},
  pages={428--436},
  year={2020},
  publisher={Nature Publishing Group UK London}
}

@online{gumin2016wfc,
  author       = {Maxim Gumin},
  title        = {Wave Function Collapse Algorithm},
  version      = {1.0},
  date         = {2016-09-30},
  url          = {https://github.com/mxgmn/WaveFunctionCollapse}
}

@inproceedings{karth2017wfc,
  title={WaveFunctionCollapse is constraint solving in the wild},
  author={Karth, Isaac and Smith, Adam M},
  booktitle={Proceedings of the 12th International Conference on the Foundations of Digital Games},
  pages={1--10},
  year={2017}
}

@book{farin2002curves,
  author = {Gerald Farin},
  title = {Curves and Surfaces for Computer-Aided Geometric Design: A Practical Guide},
  edition = {5},
  publisher = {Morgan Kaufmann},
  year = {2002}
}

@inproceedings{lavalle1998rrt,
  title={RRT-connect: An efficient approach to single-query path planning},
  author={Kuffner, James J and LaValle, Steven M},
  booktitle={Proceedings 2000 ICRA. Millennium conference. IEEE international conference on robotics and automation. Symposia proceedings (Cat. No. 00CH37065)},
  volume={2},
  pages={995--1001},
  year={2000},
  organization={IEEE}
}

@article{simba2016real,
  title={Real-time smooth trajectory generation for nonholonomic mobile robots using B{\'e}zier curves},
  author={Simba, Kenneth Renny and Uchiyama, Naoki and Sano, Shigenori},
  journal={Robotics and Computer-Integrated Manufacturing},
  volume={41},
  pages={31--42},
  year={2016},
  publisher={Elsevier}
}

@article{benko2023minimum,
  title={Minimum-time trajectory generation for wheeled mobile systems using B{\'e}zier curves with constraints on velocity, acceleration and jerk},
  author={Benko Loknar, Martina and Klan{\v{c}}ar, Gregor and Bla{\v{z}}i{\v{c}}, Sa{\v{s}}o},
  journal={Sensors},
  volume={23},
  number={4},
  pages={1982},
  year={2023},
  publisher={MDPI}
}

@inproceedings{giannousakis2018mobile,
  title={Mobile robot trajectory generation from ordered point-set using time-optimal Bezier segments},
  author={Giannousakis, Konstantinos and Tzes, Anthony},
  booktitle={2018 26th Mediterranean Conference on Control and Automation (MED)},
  pages={1--9},
  year={2018},
  organization={IEEE}
}

@article{hovrejvsi2025virtual,
  title={Virtual reality locomotion methods differentially affect spatial orientation and cybersickness during maze navigation},
  author={Ho{\v{r}}ej{\v{s}}{\'\i}, Petr and Lochmannov{\'a}, Alena and Jezl, Vojt{\v{e}}ch and Dvo{\v{r}}{\'a}k, Mat{\v{e}}j},
  journal={Scientific Reports},
  volume={15},
  number={1},
  pages={26255},
  year={2025},
  publisher={Nature Publishing Group UK London}
}

@article{bobrow1985time,
  title={Time-optimal control of robotic manipulators along specified paths},
  author={Bobrow, James E and Dubowsky, Steven and Gibson, John S},
  journal={The international journal of robotics research},
  volume={4},
  number={3},
  pages={3--17},
  year={1985},
  publisher={Sage Publications Sage CA: Thousand Oaks, CA}
}

@article{tounsi1996trajectory,
  title={Trajectory generation for mobile robots},
  author={Tounsi, Mohamed and Le Corre, JF},
  journal={Mathematics and computers in simulation},
  volume={41},
  number={3-4},
  pages={367--376},
  year={1996},
  publisher={Elsevier}
}

@article{prithul2021teleportation,
  title={Teleportation in virtual reality; a mini-review},
  author={Prithul, Aniruddha and Adhanom, Isayas Berhe and Folmer, Eelke},
  journal={Frontiers in Virtual Reality},
  volume={2},
  pages={730792},
  year={2021},
  publisher={Frontiers Media SA}
}

@article{clifton2020effects,
  title={Effects of steering locomotion and teleporting on cybersickness and presence in HMD-based virtual reality},
  author={Clifton, Jeremy and Palmisano, Stephen},
  journal={Virtual Reality},
  volume={24},
  number={3},
  pages={453--468},
  year={2020},
  publisher={Springer}
}

@incollection{powell1994direct,
  title={A direct search optimization method that models the objective and constraint functions by linear interpolation},
  author={Powell, Michael JD},
  booktitle={Advances in optimization and numerical analysis},
  pages={51--67},
  year={1994},
  publisher={Springer}
}

@article{das2025adaptable,
  title={An adaptable framework for remotely controlling a telepresence robot in a dynamic environment},
  author={Das, Barnali and Dobie, Gordon},
  journal={Robotics and Autonomous Systems},
  pages={105305},
  year={2025},
  publisher={Elsevier}
}

@inproceedings{jones2020vroom,
  title={Vroom: virtual robot overlay for online meetings},
  author={Jones, Brennan and Zhang, Yaying and Wong, Priscilla NY and Rintel, Sean},
  booktitle={Extended Abstracts of the 2020 CHI Conference on Human Factors in Computing Systems},
  pages={1--10},
  year={2020}
}

@misc{double,
    title = {Double 3},
    note = {\url{https://www.doublerobotics.com/} [Accessed: 29.6.2026]}
}

@misc{gobe,
    title = {The Future of Remote Collaboration},
    note = {\url{https://gobe.blue-ocean-robotics.com/} [Accessed: 29.6.2026]}
}

@inproceedings{beno2018work,
  title={Work flexibility, telepresence in the office for remote workers: A case study from Austria},
  author={Beno, Michal},
  booktitle={International Conference on Multi-disciplinary Trends in Artificial Intelligence},
  pages={19--31},
  year={2018},
  organization={Springer}
}

@INPROCEEDINGS{239144,
  author={Schultz, R.J. and Nakajima, R. and Nomura, J.},
  booktitle={Proceedings IECON '91: 1991 International Conference on Industrial Electronics, Control and Instrumentation}, 
  title={Telepresence mobile robot for security applications}, 
  year={1991},
  volume={},
  number={},
  pages={1063-1066 vol.2},
  doi={10.1109/IECON.1991.239144}}

@article{deitke2022️,
  title={ProcTHOR: Large-Scale Embodied AI Using Procedural Generation},
  author={Deitke, Matt and VanderBilt, Eli and Herrasti, Alvaro and Weihs, Luca and Ehsani, Kiana and Salvador, Jordi and Han, Winson and Kolve, Eric and Kembhavi, Aniruddha and Mottaghi, Roozbeh},
  journal={Advances in neural information processing systems},
  volume={35},
  pages={5982--5994},
  year={2022}
}

@article{pivtoraiko2009differentially,
  title={Differentially constrained mobile robot motion planning in state lattices},
  author={Pivtoraiko, Mihail and Knepper, Ross A and Kelly, Alonzo},
  journal={Journal of Field Robotics},
  volume={26},
  number={3},
  pages={308--333},
  year={2009},
  publisher={Wiley Online Library}
}

@article{fremont2023scenic,
  title={Scenic: A language for scenario specification and data generation},
  author={Fremont, Daniel J and Kim, Edward and Dreossi, Tommaso and Ghosh, Shromona and Yue, Xiangyu and Sangiovanni-Vincentelli, Alberto L and Seshia, Sanjit A},
  journal={Machine Learning},
  volume={112},
  number={10},
  pages={3805--3849},
  year={2023},
  publisher={Springer}
}

@inproceedings{riecke2012move,
  title={To move or not to move: can active control and user-driven motion cueing enhance self-motion perception (" vection") in virtual reality?},
  author={Riecke, Bernhard E and Feuereissen, Daniel},
  booktitle={Proceedings of the ACM symposium on applied perception},
  pages={17--24},
  year={2012}
}

@article{rouhani2024towards,
  title={Towards benchmarking VR sickness: A novel methodological framework for assessing contributing factors and mitigation strategies through rapid VR sickness induction and recovery},
  author={Rouhani, Rose and Umatheva, Narmada and Brockerhoff, Jannik and Keshavarz, Behrang and Kruijff, Ernst and Gugenheimer, Jan and Riecke, Bernhard E},
  journal={Displays},
  volume={84},
  pages={102807},
  year={2024},
  publisher={Elsevier}
}

@inproceedings{suomalainen2021comfort,
  title={Comfort and sickness while virtually aboard an autonomous telepresence robot},
  author={Suomalainen, Markku and Mimnaugh, Katherine J and Becerra, Israel and Lozano, Eliezer and Murrieta-Cid, Rafael and LaValle, Steven M},
  booktitle={International Conference on Virtual Reality and Mixed Reality},
  pages={3--24},
  year={2021},
  organization={Springer}
}

@article{westfall2015replicating,
  title={Replicating studies in which samples of participants respond to samples of stimuli},
  author={Westfall, Jacob and Judd, Charles M and Kenny, David A},
  journal={Perspectives on Psychological Science},
  volume={10},
  number={3},
  pages={390--399},
  year={2015},
  publisher={Sage Publications Sage CA: Los Angeles, CA}
}

@inproceedings{suomalainen2022unwinding,
  title={Unwinding rotations improves user comfort with immersive telepresence robots},
  author={Suomalainen, Markku and Sakcak, Basak and Widagdo, Adhi and Kalliokoski, Juho and Mimnaugh, Katherine J and Chambers, Alexis P and Ojala, Timo and LaValle, Steven M},
  booktitle={2022 17th ACM/IEEE International Conference on Human-Robot Interaction (HRI)},
  pages={511--520},
  year={2022},
  organization={IEEE}
}

@inproceedings{sakcak2019homotopy,
  title={Homotopy aware kinodynamic planning using RRT-based planners},
  author={Sakcak, Basak and Bascetta, Luca and Ferretti, Gianni},
  booktitle={2019 18th European Control Conference (ECC)},
  pages={1568--1573},
  year={2019},
  organization={IEEE}
}

@inproceedings{kuderer2014online,
  title={Online generation of homotopically distinct navigation paths},
  author={Kuderer, Markus and Sprunk, Christoph and Kretzschmar, Henrik and Burgard, Wolfram},
  booktitle={2014 IEEE International Conference on Robotics and Automation (ICRA)},
  pages={6462--6467},
  year={2014},
  organization={IEEE}
}

@article{hernandez2015comparison,
  title={A comparison of homotopic path planning algorithms for robotic applications},
  author={Hernandez, Emili and Carreras, Marc and Ridao, Pere},
  journal={Robotics and Autonomous Systems},
  volume={64},
  pages={44--58},
  year={2015},
  publisher={Elsevier}
}

\end{document}